\pdfoutput=1

\documentclass[runningheads,a4paper]{llncs}

\let\llncssubparagraph\subparagraph
\let\subparagraph\paragraph
\usepackage[compact]{titlesec}
\let\subparagraph\llncssubparagraph

\usepackage{graphicx}

\usepackage{times}
\usepackage{latexsym}
\usepackage{hyperref}

\usepackage{csquotes}

\usepackage[T1]{fontenc}

\usepackage[utf8]{inputenc}

\usepackage{microtype}
\usepackage[column=O]{cellspace}
\usepackage{booktabs}
\usepackage{adjustbox}
\usepackage{tabularx}
\usepackage[none]{hyphenat}
\usepackage{lipsum}
\usepackage{amsmath,amssymb}
\usepackage{multirow}
\usepackage{capt-of}
\usepackage[colorinlistoftodos]{todonotes}
\usepackage{caption} 
\usepackage{longtable}
\usepackage[bb=boondox]{mathalfa}
\usepackage{authblk}

\usepackage{algorithm}
\usepackage{algpseudocode}
\usepackage[
    backend=biber,
    style=nature,
  ]{biblatex}
\usepackage{titlesec}

\DeclareMathOperator*{\argmin}{arg\,min}
\algnewcommand{\LeftComment}[1]{\Statex \(\triangleright\) #1}

\title{Robust Deep Reinforcement Learning for Extractive Legal Summarization }

\author{Duy-Hung Nguyen\inst{1} \and Bao-Sinh Nguyen\inst{1} \and Nguyen Viet Dung Nghiem\inst{1} \and Dung Tien Le\inst{1} \and Mim Amina Khatun\inst{1} \and Minh-Tien Nguyen\inst{1,3} \and Hung Le\inst{2}}

\institute{Cinnamon AI \\10th floor, Geleximco building, 36 Hoang Cau, Dong Da, Hanoi, Vietnam. \\ \email{\{hector, simon, henry93, nathan, amim, ryan.nguyen\}@cinnamon.is} \and
Deakin University, Australia.\\
\email{thai.le@deakin.edu.au}\\
 \and
Hung Yen University of Technology and Education, Vietnam.\\
\email{tiennm@utehy.edu.vn}}

\authorrunning{Duy-Hung et al.}

\begin{document}
\maketitle

\begin{abstract}
Automatic summarization of legal texts is an important and still a challenging task since legal documents are often long and complicated with unusual structures and styles. Recent advances of deep models trained end-to-end with differentiable losses can well-summarize natural text, yet when applied to the legal domain, they show limited results. In this paper, we propose to use reinforcement learning to train current deep summarization models to improve their performance in the legal domain. To this end, we adopt proximal policy optimization methods and introduce novel reward functions that encourage the generation of candidate summaries satisfying both lexical and semantic criteria. We apply our method to training different summarization backbones and observe a consistent and significant performance gain across three public legal datasets. 
\keywords{Legal document  \and Summarization \and Reinforcement Learning.}

\end{abstract}

\section{Introduction}

Legal documents are long and hard to understand. These documents, either in form of court orders, contracts or terms of service, involve elaborative sentences, formal grammars and dense texts with free-flowing legal jargon  \cite{bhattacharya2019comparative,jain2021summarization}. Reading legal texts is thus challenging for both ordinary people and legal experts. 
Given the increasing number of legal texts released everyday, there is an urgent need for automating legal text summarization that is capable of shortening the original texts without losing critical information of the document.


Before the deep learning era, classical summarization methods use handcrafted features and simple statistics designed for specific type of case judgments \cite{farzindar2004letsum,polsley2016casesummarizer}. With the rise of deep learning and public legal documents being made available, there have been various attempts to train automated end-to-end legal summarization systems \cite{jain2021summarization}. One straightforward way is to  adopt powerful domain-independent summarization models, casting the problem to ranking \cite{zhong2020extractive}  or sentence classification \cite{liu2019fine}. 
All of the prior works trained these models using differentiable loss (i.e cross-entropy) to maximize the likelihood of the ground-truth summaries, showing better results compared to classical methods on some legal datasets \cite{anand2019effective,kornilova2019billsum}. 

However, the performance of general summarization methods is still limited and does not satisfy the requirement of the law industry \cite{jain2021summarization}. This is attributed to a well-known problem in text summarization \cite{narayan2018ranking}: the mismatch between the learning objective and the evaluation criterion, namely Recall-Oriented Understudy for Gisting Evaluation (ROUGE) \cite{lin2004rouge}. While the learning objective aims to maximize the likelihood of the ground-truth summary, ROUGE heavily relies on the lexical correspondence between the ground-truth and candidate summaries. The problem is even worse for the legal domain since a legal document is often very long compared to the summary, making a challenge to optimize the loss. 

\begin{figure}
    \centering
    \includegraphics[width=1\textwidth]{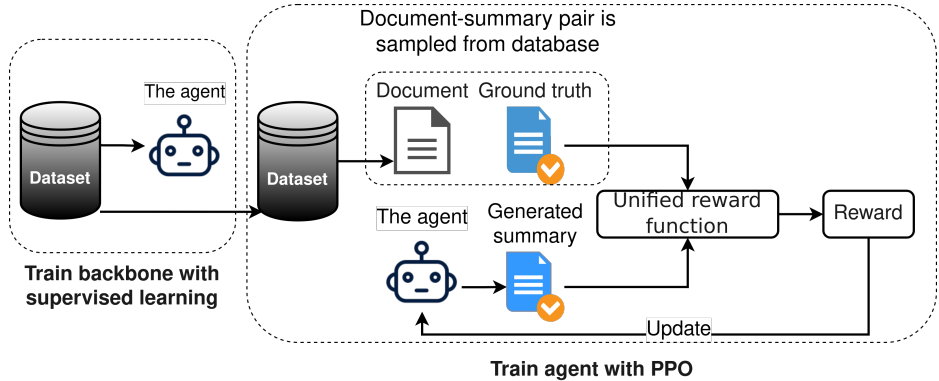}
    \caption{Our summarization framework}
    \label{fig:framework}
\end{figure}

To address this problem, we use reinforcement learning (RL) to train summarization models. By using RL, we can design complementary reward schemes that guide the learning towards objectives beyond the traditional likelihood maximization. 
Here we focus on keyword-level semantics to generate summaries which contain critical legal terms and phrases in the document. This allows our model to simulate the summarization process of humans in dealing with long legal documents. Our method is illustrated in Figure \ref{fig:framework}. We first train a backbone model to predict whether a sentence from the original document should be included in the summary using normal supervised training. Then, we finetune the model by using RL with a novel reward model that smoothly integrates lexical, sentence and keyword-level semantics into one reward function, unifying different perspectives constituting a good legal summary. 
Moreover, to ensure a stable finetuning process, we use proximal policy optimization (PPO) and enforce the exploring policy to be close to the supervised model by using Kullback-Leibler (KL) divergence as the additional intermediate rewards. 
We examine our proposed reward model on three different summarization backbones and validate the performance of our approach on three public legal datasets with different characteristics. Experimental results show that our training method consistently boosts the performance of the backbone summarization models quantitatively and qualitatively. 



Our contributions are three-fold: (1) we pioneer employing reinforcement learning (RL) to train summarization models for the legal domain; (2) we construct a new training objective in form of RL rewards to facilitate both semantics and lexical requirements; (3) we evaluate the proposed method on diverse legal document types. The proposed method gains significant improvement over other baselines across 3 datasets.

\section{Background}

\subsection{Problem formulation}
We formulate the problem as extractive summarization. Given a document $D$ with $n$ sentences $D = \{s_1, s_2, …, s_n\}$, the model extracts $m$ salient sentences (where $m < n$) and re-organizes them as a summary $S$. In most cases, the summarization model can be viewed as a binary classification task, where it predicts a label $y_i \in \{0, 1\}$ for each sentence (a $y_i = 1$ label denotes that the $i$-th sentence should be included in the summary). To do this, the model learns to assign a score $\pi(y_i|s_i,D,\theta)$ that quantifies the importance of each sentence, where $\theta$ is the learned parameter of a neural network. After training, the model makes prediction to select top $m$ sentences with the highest scores among $\pi(y_i=1|s_i,D,\theta)$ as a summary.

\subsection{Pitfalls of current reinforcement training for summarization}


To train the summarization model with RL, prior works normally use a simple  policy gradient method-REINFORCE as the optimization algorithm and directly adopts ROUGE scores as a global reward. In this section, we give a background on these traditional methods and point out their limitations.



\paragraph{\textbf{REINFORCE}}
\cite{williams1992simple} relies on direct differentiation of the RL objective, which is the expected total reward over time $J(\theta) = \mathbb{E}_{\tau \sim p(\tau)}{[\sum_{i=1}^{n}r_i(s_i, y_i)]}
$, where $\tau$ denotes sampled trajectory, $s_i$ is the state of the agent, and $y_i$ is the action taken by the agent at the time step $i$. The key idea of $J(\theta)$ is to reinforce good actions to push up the probabilities of actions that lead to a higher total reward, and push down the probabilities of actions that lead to a lower total reward, until the model obtains an optimal policy.


The updated gradient in REINFORCE is too sensitive to the choice of learning rate. If parameters are updated with large values at a learning step, this can lead to a large change of the policy. Otherwise, too small learning rate can hopelessly lead to a slow learning progress. This might affect the summarization because summaries collected from a bad policy will guide the learning and gradually makes the policy to be far from the optimal solution.

 


\paragraph{\textbf{ROUGE as a reward}}
ROUGE-scores \cite{lin2004rouge} are often used as the reward for reinforcement training of summarization models \cite{narayan2018ranking} such as $R_{ROUGE} = \frac{R-1 + R-2 + R-L}{3}$, where $R$ denotes  the F1-scores of the corresponding ROUGE. However, ROUGE does not assess the fluency of the summary. It only tries to assess the adequacy, by simply counting $n$-grams overlapping between the extracted summary and the gold summary. Moreover, $n$-grams suffer from potential synonym or equivalent phrases appeared in a summary.



\section{Summarization with the Unified Reward}

\subsection{Supervised summarization backbone model}

\paragraph{\textbf{Document encoder}} 

We follow BERTSUM \cite{liu2019text} to encode input sentences of a document. To do that, a special [CLS] token (short for "classification") was inserted at the start position of every sentence. We also modify the interval segment embeddings by assigning embedding $E_A$ or $E_B$ for the i-th sentence, depending on its odd or even position. This way, both tokens’ relations and inter-relations among sentences are learned simultaneously through different layers of the Transformer encoder.
The output of BERT returns representations of each token, where the embedding of the i-th [CLS] token from the top layer of the Transformer encoder (denoted as $h_i^L$) is used as the sentence representation of $s_i$.

\paragraph{\textbf{Sentence selector}}
Given the sentence representation, neural networks are used to predict whether to select the sentence as part of the summary. We examine with three architectures: linear, LSTM, and Transformer to create three different summarization backbones. The output of each backbone is followed by a sigmoid function to compute the score for each sentence as follows.

\begin{itemize}
    \item a linear layer:
        \begin{equation} \label{eq:linear_layer}
            y_i = sigmoid(W_0 h_i^L + b_0)
        \end{equation}
    \item an LSTM layer \cite{hochreiter1997long}:
        \begin{equation} \label{eq:lstm_layer_1}
            u_i = \overrightarrow{\text{LSTM}}(h_i^L)
        \end{equation}
        \begin{equation} \label{eq:lstm_layer_2}
            y_i = sigmoid(W_0 u_i + b_0)
        \end{equation}
    \item a Transformer encoder block with multi-head attention followed by two fully-connected feed-forward network (FFN) \cite{vaswani2017attention}. The attention is:
    \begin{equation}
        Attention(Q, K, V) = softmax\left(\frac{QK^T}{\sqrt{d_k}}\right)V
    \end{equation}
where $Q$, $K$, $V$ are matrices from the embeddings of all tokens in the i-th sentence.
        
\end{itemize}

\subsection{Finetuning with Proximal Policy Optimization}\label{sec:PPO}

\paragraph{\textbf{PPO}}

After training the backbone model by using the standard cross-entropy loss, we consider the trained backbone model as the initial policy, and then continue finetuning the policy with RL. Here we choose Proximal policy optimization (PPO) \cite{schulman2017proximal} as our RL algorithm to optimize the policy because it works well as optimization algorithm in NLP domain \cite{stiennon2020learning,ziegler2019fine}. PPO controls the change in the policy being updated at each iteration, so that the policy does not move too far. We hypothesize that it can give benefit to our policy for extracting better summaries.

\paragraph{\textbf{KL reward}}
We follow \cite{ziegler2019fine,stiennon2020learning} to define the reward scheme for RL. Let $\pi^\mathrm{SL}$ denotes the supervised trained backbone model and $\pi^\mathrm{RL}$ the one that we optimize with RL. We calculate the reward at the time step $i$ as:

\begin{equation}
    r_i = -\beta_{KL}*\log{[\frac{\pi^{\mathrm{RL}} (y_i|s_i, D, \theta)}{\pi^{\mathrm{SL}} (y_i|s_i, D, \theta)}]} + \mathbb{1}(i=n) \times R_\mathrm{unified}(S)
\end{equation}
where $\beta_{KL}$ is the KL coefficient, $n$ is the final time step corresponding to the total number of sentences in the document $D$.

For the intermediate time step $i$ ($i < n$), the reward is just the negative KL divergence between the output's distribution of the backbone model and the current policy. It ensures to prevent the current policy from generating outputs which are too different from the outputs of the backbone model. For the final time step $i$ ($i = n$), when the model obtains the entire summary $S$, a reward term $R_\mathrm{unified}(S)$ that we design to measure the quality of the extracted summary as a whole. We introduce our reward in Section \ref{sec:our-reward}.

\subsection{The unified reward}\label{sec:our-reward}

As mentioned, the reward using ROUGE-scores only considers $n$-grams overlapping between an extracted summary and the ground-truth reference. It ignores the semantic aspect of word and sentence levels. We argue that the reward function should encode the semantics to guide the summarization model to output a good summary, that can reach human level. To exploit the semantic aspect, we introduce a unified reward function in Eq. \eqref{eq:ru}, 
which considers different important aspects of a good summary.

\begin{equation}\label{eq:ru}
R_{unified} = \alpha_1 R_{ROUGE} + \alpha_2 R_{kw}   + \alpha_3 R_{seq} 
\end{equation}

where $\alpha_1$, $\alpha_2$, and $\alpha_3$ are the control coefficients; $R_{ROUGE}$ is the ROUGE-score function; $R_{kw}$ considers the keyword semantic; and $R_{seq}$ captures the semantics of a sequence. The new reward function includes three components. The ROUGE function encodes the word overlapping between an extracted summary and the gold reference. We use this function to directly force the backbone model to extract important sentences which tend to be similar to the gold reference \cite{narayan2018ranking}. The $R_{kw}$ supports the ROUGE function in term of semantics. This is because the ROUGE function only considers the $n$-grams overlapping aspect. In many cases, words in an extracted summary and the gold reference are different in term of characters, but they share similar meaning. Therefore, we design the $R_{kw}$ function to address this problem. Finally, the $R_{seq}$ helps the backbone model to extract sequences similar to the target text.

\paragraph{\textbf{The $R_{kw}$ function}}
To produce $R_{kw}$, we first use BERT to embed phrases and utilize the cosine function to compute the similarity among embedded vectors. 
Then, the method shown in Algorithm \ref{algo:kw} is used to produce two sets of keywords $\mathbb{K}_D, \mathbb{K}_S$ from the original document $D$ and the summary $S$. For each keyword in $\mathbb{K}_D$, we find the most similar keywords in $\mathbb{K}_S$. Finally, the keyword reward is computed by the average of all similarities.
\begin{equation}
R_{kw} = \frac{\sum_{kw \in \mathbb{K}_D} \max_{kw' \in \mathbb{K}_S}{consine(kw,kw')}}{|\mathbb{K}_D|}
\end{equation}
\begin{algorithm}[t]
\caption{Algorithm for Get\_Keywords.}
\label{algo:kw}
\begin{algorithmic}[1]
\Require {A document: $D$ and number of keywords $n_k$;}
\Ensure {Keyword set $\mathbb{K}$;}
\State $\mathbb{S} \leftarrow$ Top $2*n_k$ keywords from $D$ {\(\triangleright\)\scriptsize\it Select based on $ \|emb(phrase)\|$}
\State $\mathbb{K} \leftarrow \{\mathbb{S}[1]\}$  {\(\triangleright\)\scriptsize\it Take the first element of $\mathbb{S}$}
\While {$|\mathbb{K}| < n_k$}
    \State $best\_kw \leftarrow \argmin_{kw \in \mathbb{S} \setminus \mathbb{K}} \left( \max_{kw' \in \mathbb{K}} cosine(kw,kw') \right)$ 
    \Statex \hspace{1cm}{\(\triangleright\) \scriptsize\it Find the least similarity keyword to the current keyword set $\mathbb{K}$}
    \State $\mathbb{K} \leftarrow \mathbb{K} \cup \{best\_kw\} $
\EndWhile
\State Return $\mathbb{K}$;
\end{algorithmic}
\end{algorithm}

Opposed to DSR \cite{li2019deep}, we select the list of keywords before comparing them.  It is essential for legal documents as keywords take a vital role in the text's meaning and amplify their effect on the reward function. In addition, several terms used in the legal domain can have the same meaning. Therefore, to reduce the redundancy and inaccuracy, we take a phrase that is the most different from the current set of keywords when choosing keywords. In practice, the size of the keyword set $n_k$ is set as 3 in all experiments.


\paragraph{\textbf{The semantic sentence function}}
Since $R_{kw}$ only encourages similar keywords appearing in the summary, it does not guarantee a the coherence of the whole summary.
Thus, we define $R_{seq}$ to enforce the semantic similarity between the final summary $S$ and the gold reference $S*$. 
Initially, \texttt{Moverscore} \cite{kusner2015word} on Word2Vec \cite{mikolov2013linguistic} is used to measure the translating distance between the two texts $d(S, S*)$. Concretely, the reward is defined as $R_{seq} = \frac{1}{d(S, S*)+\epsilon}$ where $\epsilon=10^{-2}$. 

\subsection{Training and inference}

The backbone model was trained to to initialize the policy. Due to the limitation of resources, the model bases on BERT-base with 12 Transformer blocks, the hidden size of 768, and 110M parameters. During RL training, the PPO algorithm was adopted to optimize the policy. The Adam optimizer \cite{kingma2014adam} was used to optimize the reward function in Eq. \ref{eq:ru} with the learning rate of 1e-5. At the test time, the model selects top $m$ sentences that have the highest probabilities predicted by the trained policy to form a summary.
 
\section{Experimental Setup}

\subsection{Datasets} 

We used three legal datasets for evaluation as follows.

\subsubsection{Plain English Summarization of Contracts (PESC)} is a legal dataset written in plain English, in which a text snippet is paired with an abstractive summary \cite{manor2019plain}. Each legal snippet contains approximately 595 characters. The length of a summary is around 202 characters. 

\subsubsection{BillSum} consists of 22,218 US Congressional bills with human-written reference summaries collected from United  States  Government Publishing Office \cite{kornilova2019billsum}. The data was split into 18,949 training bills and 3,269 testing bills. Each document contains around 46 sentences and each summary includes around 6 sentences.

\subsubsection{Legal Case Reports dataset (LCR)} contains 3,890 Australian legal cases from the Federal Court of Australia \cite{galgani2012combining}. Each case contains around 221 sentences and 8.3 catchphrases which is considered as a summary. We divided the data into three sets: training (2,590), validation (800), and testing (500 samples).

\subsection{Settings and evaluation metrics}
The summarization model is based on BERT-base with 12 Transformer blocks, the hidden size of $h_i^L$ from BERT is 768. We used one layer for the backbone of LSTM and two linear layers for the backbone of transformers with the output vector of 768. All models were trained with a batch size of 256 on a single Tesla T4 GPU. In our experiment, we set $\alpha_1=0.3$, $\alpha_2=0.4$, and $\alpha_3=0.3$ for $R_{unified}$ by using random search tuning on the validation set. We set $\beta_{KL}=0.05$ for the KL coefficient.
At the testing time, the number of selected sentences was set as 3 for PESC, 6 for BillSum, and 8 for LCR datasets. We used ROUGE-scores\footnote{The parameter: -c 95 -m -r 1000 -n 2.} to measure the quality of summaries. We report the F-score of ROUGE-1, ROUGE-2, and ROUGE-L as the main metrics.

\section{Experimental Results}

\subsection{Legal Cases: Working with different summarization backbones}

In this section, we report quantitative improvement over the backbone baselines on 2 formal and complicated legal datasets: Billsum and LCR. Here,
\textbf{Backbone} is the BERTSum model with three different sentence selectors: Linear, LSTM, and Transformer. 
 In addition, we also compare our backbone to other classical legal summarization baselines: CaseSummarizer \cite{polsley2016casesummarizer} and Restricted Boltzmann Machine (RBM) \cite{verma2018extractive}.
 \begin{table}[t]
 \begin{center}
\small
\begin{tabular*}{\linewidth}{@{\extracolsep{\fill}} l | c c c | c c c }
    \hline
    \multirow{2}{*}{Model} & 
    \multicolumn{3}{c}{BillSum} &
    \multicolumn{3}{c}{LCR} \\ \cline{2-7}
    & ROUGE-1 & ROUGE-2 & ROUGE-L & ROUGE-1 & ROUGE-2 & ROUGE-L \\
    \hline
    CaseSummarizer \cite{polsley2016casesummarizer} & 27.32 & 15.65 & 25.98 & 1.49 & 0.59 & 1.31 \\
    RBM \cite{verma2018extractive} & 27.86 & 13.58 & 25.18 & 1.66  & 0.83 & 1.44  \\
    \hline
    \multicolumn{7}{l}{\textbf{The Backbone}} \\
    \hline
    Linear & 31.87 & 11.46 & 28.77 & 21.79 & 5.96 & 20.42 \\
    LSTM & 33.56 & 12.67 & 30.42 & 22.76 & 6.34 & 21.29 \\
    Transformer & 32.58 & 11.68 & 29.44 & 23.09 & 6.59 & 21.59 \\
    \hline
    \multicolumn{7}{l}{\textbf{RL with $R_{unified}$}} \\
    \hline
    Linear+$R_{unified}$  & 33.31 & 12.62 & 30.30 & 21.81 & 6.00 & 20.43\\
    LSTM+$R_{unified}$ & \textbf{33.57} & 12.68 & \textbf{30.42} & 22.8 & 6.33 & 21.31 \\
    Transformer +$R_{unified}$ & 33.33 & \textbf{12.70} & 30.21 & \textbf{23.21} & \textbf{6.92} & \textbf{21.72} \\
    \hline
\end{tabular*}
\captionof{table}{\label{table:backbones}Evaluation scores for all backbones.}
\end{center}
\end{table}

Table \ref{table:backbones} reports the ROUGE scores of the backbone before and after training with RL. Notably, our backbones achieve strong results, significantly outperforming traditional legal summarizers. In this case, classical methods use handcrafted features and only work for certain types of legal documents, thus showing poor results on the datasets.
 We observe that after training with our reward function, the performance of our summarization backbones gets improvement by a substantial amount. 
 The improvements indicates that the proposed method can effectively improve all types of the pretrained-supervised model. The new reward function in Eq. (\ref{eq:ru}) forces the summarization model to select salient sentences which are good in term of lexical and semantic aspects on the word and sentence levels. Overall, the Transformer backbone gives the best results. The possible reason is that this backbone uses a more complicated architecture than Linear and LSTM. Due to the high accuracy, we use the Transformer as the main backbone in other experiments. 



\subsection{Terms of Service: State-of-the-art results on PESC}


In this section, we compare our method to strong methods from the literature on PESC dataset. \textbf{Lead-$m$} extracts $m$ first sentences of a document to form a summary \cite{manor2019plain}.
\textbf{ROUGE} \cite{lin2004rouge} uses the ROUGE-L as the final reward to optimize the policy \cite{li2019deep,pasunuru2018multi}.
\textbf{DSR} \cite{li2019deep} is a strong reward method that boost the performance of backbone \cite{li2019deep}. 
\textbf{BLANC} \cite{vasilyev2020fill} is also a metric to estimate the quality of a summary. This metric will be used to provide a reward signal when optimizing the policy.
\textbf{REFRESH} \cite{narayan2018ranking} uses an encoder-decoder to rank sentences, then the top-ranked sentences are assembled as a summary. The model was updated using REINFORCE algorithm.

Table \ref{table:rewards} describes the best results on the PESC dataset. For reward-based methods, all models in this experiment are in the same configuration except for the final reward. It is noticeable that all these models improve the supervised backbone (Transformer). 
Our model with $R_{unified}$ achieves SOTA result on the dataset. There is also a large performance gap between our method and baselines: REFRESH and Lead-$m$. Our method achieves 2.23\%, 1.3\%, and 1.45\% higher than REFRESH on ROUGE-1, ROUGE-2, and ROUGE-L respectively. In comparison with Lead-$m$, which is the best model reported in \cite{manor2019plain}, our approach performs nearly 5 ROUGE-L points higher.

\begin{table}[t]
\begin{center}
\small
\begin{tabular*}{\linewidth}{@{\extracolsep{\fill}} l | c c c}
\hline
Model & ROUGE-1 & ROUGE-2 & ROUGE-L \\
\hline
Lead-$m$ \cite{manor2019plain} & 24.38 & 7.52 & 17.63  \\
REFRESH \cite{narayan2018ranking} & 23.47 & 8.23 & 20.83 \\
Backbone & 24.67 & 8.91 & 20.76 \\
ROUGE \cite{lin2004rouge} & 25.52 & 9.50 & 21.50 \\
DSR \cite{li2019deep} & 25.53 & 9.35 & 21.76 \\
BLANC \cite{vasilyev2020fill} & 25.56 & 9.51 & 21.96 \\
$R_{unified}$ & \textbf{25.70} & \textbf{9.53} & \textbf{22.30} \\
\hline
\end{tabular*}
\captionof{table}{\label{table:rewards}Best results on PESC
dataset.}
\end{center}
\end{table}

\subsection{Ablation studies}

\begin{table}[t]
\begin{center}
\small
\setlength\tabcolsep{0pt}
\begin{tabular*}{\linewidth}{@{\extracolsep{\fill}} l | c c c } \hline
 Model & ROUGE-1 & ROUGE-2 & ROUGE-L \\
\hline
Ours w/o KL & 24.83 & 8.95 & 21.36 \\
Ours w/o PPO & 24.79 & 9.13 & 21.17 \\
Ours w/o $R_{kw}$ & 25.65 & 9.52 & 21.92 \\
Ours w/o $R_{seq}$ & 25.61 & 9.48 & 21.85 \\
Ours full & \textbf{25.70} & \textbf{9.53} & \textbf{22.3} \\ \hline
\end{tabular*}
\captionof{table}{\label{tab:pesc abl_1} Effect of ablating components of our model on PESC dataset.}
\end{center}
\end{table}


We conduct more evaluations to assess the impact of KL reward and PPO on the proposed method. 
We also validate components of our $R_{unified}$ including $R_{kw}$ and $R_{seq}$. Table \ref{tab:pesc abl_1} shows that the proposed approach obtains significant benefits from KL rewards. Without the KL rewards, ROUGE scores drop approximately 1\%. We also observe similar drops in performance when we replace PPO with the classical REINFORCE, which stresses the importance of using proximal policy optimization.  Training without $R_{seq}$ and $R_{kw}$ also reduces the performance of the baselines. Since ROUGE-score is our objective, combining $R_{ROUGE}$ with our $R_{seq}$ and $R_{kw}$ will provide a sufficient training signal to the backbone. As expected, the $R_{unified}$ performs best.

\subsection{A case study}
In this section, we provide a closer look at the extracted summary produced by RL and compare it to the output of supervised-trained model (SL summary), and the gold summary.

\begin{longtable}{|p{1\linewidth}|}
\hline
 \textbf{Catchpharases (gold summary):} appeal from a decision of the federal magistrates court . application for a protection visa . serious personal assault . whether tribunal failed to comply with s 430 of the migration act . whether tribunal failed to take adequately into account relevant material . whether tribunal failed adequately to take into account integer of \textcolor{red}{appellant} husband 's claims . reviewable error established . migration\\ \hline
 \textbf{SL summary:} the tribunal 's decision 3 the \textcolor{red}{appellant} husband provided to the tribunal a comprehensive statutory declaration of 69 paragraphs assembled on 27 october 2004 , \textbf{...}
 whereby his honour dismissed the \textcolor{red}{appellants} ' application for judicial review of the decision of the refugee review tribunal ( ` the tribunal ' ) \textbf{...}
 he had enjoyed an association with the leadership of the united national party ( ` unp ' ) in sri lanka , being an association which had largely formed the political context to the present claims to refugee status of the \textcolor{red}{appellants} . \textbf{...}
 \\ \hline
 \textbf{RL summary:} \textcolor{blue}{the second \textcolor{red}{appellant} is the wife of the first \textcolor{red}{appellant} 's second marriage which took place in 1988 , and both of the \textcolor{red}{appellants} are of sinhalese ethnicity .} the tribunal 's decision 3 the \textcolor{red}{appellant} husband provided to the tribunal a comprehensive statutory declaration of 69 paragraphs assembled on 27 october 2004 , \textbf{...}
 whereby his honour dismissed the \textcolor{red}{appellants} ' application for judicial review of the decision of the refugee review tribunal ( ` the tribunal ' ) made on 11 april 2005 and handed down on 4 may 2005 . \textbf{...}
 \\ \hline
\caption{A case study in LCR dataset.}
\label{tab:casestudy_lcr}
\end{longtable}

Table \ref{tab:casestudy_lcr} shows an example from the LCR dataset. Thanks to the keyword-level semantic reward, the important term "appellant(s)" (highlighted with red color) existed in the gold summary is presented more times in the RL summary. The KL reward also ensures that the RL summary is not too different from the SL summary. The only different sentence in the RL summary compared to the SL summary is highlighted with the blue color, which gives readers a more clear context for this particular legal case.

\section{Related Works}


Recently, there have been large amounts of studies on document summarization using reinforcement learning. For this task, most of the research employs discrete functions like ROUGE as a part of reward \cite{wu2018learning,li2019deep}. \cite{wu2018learning} trained a model to evaluate the coherence of current text and used the coherent score as the intermediate reward. In contrast to the original ROUGE score, which treats all words in the text equally, \cite{pasunuru2018multi} introduced a novel salient reward that gives high weight to the critical words in summary, and an entailment scorer gives a high reward to the logically entailed summaries. The most similar to ours is \cite{li2019deep}. The authors proposed distributional semantic reward to capture semantic relation between similar words. However, their study focus on the semantic of the entire text. In contrast, we try to simulate the summarization process of legal experts, so our research introduces a reward that focuses on keyword-level semantics.

In the legal area, many approaches for text summarizing have been presented. FLEXICON\cite{gelbart1991flexicon} is the pioneer in this field. It references keywords of original text against a large database to find a candidate summary. \cite{kim2012summarization} considered a document as a weighted graph, in which sentences were represented as nodes of the graph. Then the summary was a collection of high node values. \cite{galgani2012combining} introduced a rule-based system using manual knowledge acquisition which combines different summarization techniques. Recently, \cite{pandya2019automatic} proposed to combine K-mean clustering and TF-IDF word vectorizer to summarize legal case reports. However, to the best of our knowledge, the adaptation of reinforcement learning to legal document summarization is still an open question.

\section{Conclusion}

In this paper, we have presented a new method for summarizing various types of legal documents. 

Our approach for training the summarization model with a novel reward scheme has reached a ROUGE-score of 25.70\% for the PESC dataset, achieving the SOTA result. To further validate our approach, we extensively experiment on different datasets with different configurations. Experimental results show that the method  is consistently better than the strong baselines on additional BillSum and Legal Case Reports datasets.

Future work will expand our research to tasks such as case entailment: using reinforcement learning with novel reward designs to find precedents from the law database given a query case to achieve expert-level results.

\printbibliography

@inproceedings{wu2018learning,
  title={Learning to extract coherent summary via deep reinforcement learning},
  author={Wu, Yuxiang and Hu, Baotian},
  booktitle={Proceedings of the AAAI Conference on Artificial Intelligence},
  volume={32},
  number={1},
  year={2018}
}

@article{pasunuru2018multi,
  title={Multi-reward reinforced summarization with saliency and entailment},
  author={Pasunuru, Ramakanth and Bansal, Mohit},
  journal={arXiv preprint arXiv:1804.06451},
  year={2018}
}

@inproceedings{kim2012summarization,
  title={Summarization of legal texts with high cohesion and automatic compression rate},
  author={Kim, Mi-Young and Xu, Ying and Goebel, Randy},
  booktitle={JSAI International Symposium on Artificial Intelligence},
  pages={190--204},
  year={2012},
  organization={Springer}
}

@inproceedings{polsley2016casesummarizer,
  title={Casesummarizer: A system for automated summarization of legal texts},
  author={Polsley, Seth and Jhunjhunwala, Pooja and Huang, Ruihong},
  booktitle={Proceedings of COLING 2016, the 26th international conference on Computational Linguistics: System Demonstrations},
  pages={258--262},
  year={2016}
}

@article{pandya2019automatic,
  title={Automatic Text Summarization of Legal Cases: A Hybrid Approach},
  author={Pandya, Varun},
  journal={arXiv preprint arXiv:1908.09119},
  year={2019}
}

@inproceedings{galgani2012combining,
  title={Combining different summarization techniques for legal text},
  author={Galgani, Filippo and Compton, Paul and Hoffmann, Achim},
  booktitle={Proceedings of the workshop on innovative hybrid approaches to the processing of textual data},
  pages={115--123},
  year={2012}
}

@inproceedings{lin2004rouge,
  title={Rouge: A package for automatic evaluation of summaries},
  author={Lin, Chin-Yew},
  booktitle={Text summarization branches out},
  pages={74--81},
  year={2004}
}

@article{li2019deep,
  title={Deep reinforcement learning with distributional semantic rewards for abstractive summarization},
  author={Li, Siyao and Lei, Deren and Qin, Pengda and Wang, William Yang},
  journal={arXiv preprint arXiv:1909.00141},
  year={2019}
}

@article{vasilyev2020fill,
  title={Fill in the blanc: Human-free quality estimation of document summaries},
  author={Vasilyev, Oleg and Dharnidharka, Vedant and Bohannon, John},
  journal={arXiv preprint arXiv:2002.09836},
  year={2020}
}

@article{williams1992simple,
  title={Simple statistical gradient-following algorithms for connectionist reinforcement learning},
  author={Williams, Ronald J},
  journal={Machine learning},
  volume={8},
  number={3-4},
  pages={229--256},
  year={1992},
  publisher={Springer}
}

@article{liu2019text,
  title={Text Summarization with Pretrained Encoders},
  author={Liu, Yang and Lapata, Mirella},
  journal={arXiv preprint arXiv:1908.08345v2},
  year={2019}
}

@article{schulman2017proximal,
  title={Proximal policy optimization algorithms},
  author={Schulman, John and Wolski, Filip and Dhariwal, Prafulla and Radford, Alec and Klimov, Oleg},
  journal={arXiv preprint arXiv:1707.06347},
  year={2017}
}

@inproceedings{vaswani2017attention,
  title={Attention Is All You Need},
  author={Vaswani, Ashish and Shazeer, Noam and Parmar, Niki and Uszkoreit, Jakob and Jones, Llion and Gomez, Aidan N. and Kaiser, Łukasz and Polosukhin, Illia},
  booktitle={Proceedings of the 31st Conference on Neural Information Processing Systems},
  pages={6000--6010},
  year={2017}
}

@article{stiennon2020learning,
  title={Learning to summarize from human feedback},
  author={Stiennon, Nisan and Ouyang, Long and Wu, Jeff and Ziegler, Daniel M and Lowe, Ryan and Voss, Chelsea and Radford, Alec and Amodei, Dario and Christiano, Paul},
  journal={arXiv preprint arXiv:2009.01325},
  year={2020}
}

@article{ziegler2019fine,
  title={Fine-tuning language models from human preferences},
  author={Ziegler, Daniel M and Stiennon, Nisan and Wu, Jeffrey and Brown, Tom B and Radford, Alec and Amodei, Dario and Christiano, Paul and Irving, Geoffrey},
  journal={arXiv preprint arXiv:1909.08593},
  year={2019}
}

@article{manor2019plain,
  title={Plain english summarization of contracts},
  author={Manor, Laura and Li, Junyi Jessy},
  journal={arXiv preprint arXiv:1906.00424},
  year={2019}
}

@article{kornilova2019billsum,
  title={BillSum: a corpus for automatic summarization of US legislation},
  author={Kornilova, Anastassia and Eidelman, Vlad},
  journal={arXiv preprint arXiv:1910.00523},
  year={2019}
}

@article{hochreiter1997long,
  title={Long short-term memory},
  author={Hochreiter, Sepp and Schmidhuber, J{\"u}rgen},
  journal={Neural computation},
  volume={9},
  number={8},
  pages={1735--1780},
  year={1997},
  publisher={MIT Press}
}

@inproceedings{bhattacharya2019comparative,
  title={A comparative study of summarization algorithms applied to legal case judgments},
  author={Bhattacharya, Paheli and Hiware, Kaustubh and Rajgaria, Subham and Pochhi, Nilay and Ghosh, Kripabandhu and Ghosh, Saptarshi},
  booktitle={European Conference on Information Retrieval},
  pages={413--428},
  year={2019},
  organization={Springer}
}

@article{jain2021summarization,
  title={Summarization of legal documents: Where are we now and the way forward},
  author={Jain, Deepali and Borah, Malaya Dutta and Biswas, Anupam},
  journal={Computer Science Review},
  volume={40},
  pages={100388},
  year={2021},
  publisher={Elsevier}
}

@inproceedings{farzindar2004letsum,
  title={LetSum, an Automatic Text Summarization system in Law field},
  author={Farzindar, Atefeh and Lapalme, Guy},
  year={2004},
  organization={Legal Knowledge and Information Systems 17 th Annual Conference, Jurix}
}

@inproceedings{zhong2020extractive,
  title={Extractive Summarization as Text Matching},
  author={Zhong, Ming and Liu, Pengfei and Chen, Yiran and Wang, Danqing and Qiu, Xipeng and Huang, Xuan-Jing},
  booktitle={Proceedings of the 58th Annual Meeting of the Association for Computational Linguistics},
  pages={6197--6208},
  year={2020}
}

@article{liu2019fine,
  title={Fine-tune BERT for extractive summarization},
  author={Liu, Yang},
  journal={arXiv preprint arXiv:1903.10318},
  year={2019}
}

@article{anand2019effective,
  title={Effective deep learning approaches for summarization of legal texts},
  author={Anand, Deepa and Wagh, Rupali},
  journal={Journal of King Saud University-Computer and Information Sciences},
  year={2019},
  publisher={Elsevier}
}

@inproceedings{narayan2018ranking,
  title={Ranking Sentences for Extractive Summarization with Reinforcement Learning},
  author={Narayan, Shashi and Cohen, Shay B and Lapata, Mirella},
  booktitle={Proceedings of the 2018 Conference of the North American Chapter of the Association for Computational Linguistics: Human Language Technologies, Volume 1 (Long Papers)},
  pages={1747--1759},
  year={2018}
}

@article{kingma2014adam,
  title={Adam: A method for stochastic optimization},
  author={Kingma, Diederik P and Ba, Jimmy},
  journal={arXiv preprint arXiv:1412.6980},
  year={2014}
}

@article{gelbart1991flexicon,
  title={Flexicon, a new legal information retrieval system},
  author={Gelbart, Daphne and Smith, JC},
  journal={Can. L. Libr.},
  volume={16},
  pages={9},
  year={1991},
  publisher={HeinOnline}
}

@inproceedings{mikolov2013linguistic,
  title={Linguistic regularities in continuous space word representations},
  author={Mikolov, Tom{\'a}{\v{s}} and Yih, Wen-tau and Zweig, Geoffrey},
  booktitle={Proceedings of the 2013 conference of the north american chapter of the association for computational linguistics: Human language technologies},
  pages={746--751},
  year={2013}
}

@inproceedings{kusner2015word,
  title={From word embeddings to document distances},
  author={Kusner, Matt and Sun, Yu and Kolkin, Nicholas and Weinberger, Kilian},
  booktitle={International conference on machine learning},
  pages={957--966},
  year={2015},
  organization={PMLR}
}

@article{verma2018extractive,
  title={Extractive Summarization using Deep Learning},
  author={Verma, Sukriti and Nidhi, Vagisha},
  journal={Research in Computing Science},
  volume={147},
  pages={107--117},
  year={2018}
}
\end{document}